\documentclass[runningheads]{llncs}
\usepackage{graphicx}
\usepackage{algorithm}
\usepackage{algpseudocode}
\usepackage{array}
\usepackage[inline]{enumitem}
\usepackage{hyperref}
\usepackage{csquotes}
\usepackage{subcaption}
\usepackage{xspace}
\newcommand{\Ie}{\emph{i.e.},\xspace}
\newcommand{\Eg}{\emph{e.g.}:\xspace}

\newcommand{\Rone}{\mathbf{R_1}}
\newcommand{\Rtwo}{\mathbf{R_2}}

\begin{document}
\title{Exploring Rawlsian Fairness for \\ K-Means Clustering}
%
%
\author{Stanley Simoes \hspace{0.2in}
Deepak P \hspace{0.2in}
Muiris MacCarthaigh}
\authorrunning{S. Simoes et al.}
%
\institute{Queen's University Belfast, UK\\
\email{ssimoes01@qub.ac.uk} \hspace{0.1in} \email{deepaksp@acm.org} \hspace{0.1in} \email{M.MacCarthaigh@qub.ac.uk}}


%
\maketitle              
\begin{abstract}
We conduct an exploratory study that looks at incorporating John Rawls' ideas on fairness into existing unsupervised machine learning algorithms.
Our focus is on the task of \emph{clustering}, specifically the \emph{k-means clustering} algorithm.
To the best of our knowledge, this is the first work that uses Rawlsian ideas in clustering.
Towards this, we attempt to develop a \emph{postprocessing} technique \Ie one that operates on the cluster assignment generated by the standard k-means clustering algorithm.
Our technique perturbs this assignment over a number of iterations to make it fairer according to Rawls' \emph{difference principle} while minimally affecting the overall utility.
As the first step, we consider two simple perturbation operators -- $\Rone$ and $\Rtwo$ -- that reassign examples in a given cluster assignment to new clusters; $\Rone$ assigning a single example to a new cluster, and $\Rtwo$ a pair of examples to new clusters.
Our experiments on a sample of the Adult dataset demonstrate that both operators make meaningful perturbations in the cluster assignment towards incorporating Rawls' difference principle, with $\Rtwo$ being more efficient than $\Rone$ in terms of the number of iterations.
However, we observe that there is still a need to design operators that make significantly better perturbations.
Nevertheless, both operators provide good baselines for designing and comparing any future operator, and we hope our findings would aid future work in this direction.

\end{abstract}
\section{Introduction}
While traditional machine learning (ML) algorithms aim to maximise utility through discrimination, they are now being increasingly regulated by law so as to prevent any unjustifiable discrimination in the society, especially when these algorithms are deployed on a large scale and can heavily influence people's life chances.
These laws provide the interpretation of fairness (\Eg the \emph{four-fifths rule}\footnote{\url{https://www.law.cornell.edu/cfr/text/29/1607.4}} in the US and the \emph{Equality Act 2010}\footnote{\url{https://www.legislation.gov.uk/ukpga/2010/15/contents}} in the UK) which existing algorithms are required to comply with.
The challenge lies in translating these legal interpretations of fairness, which are in natural language, to mathematical formulations that can be incorporated in ML algorithms with minimal impact on the utility of these algorithms.

Among the many ideas of justice and fairness put forth by political philosophers and social scientists over time, the ML research community has looked at the idea of fairness that is computationally easy to model such as individual fairness \cite{dwork12fairness} and group fairness \cite{pedreshi08discrimination}.
However, this idea does not have enough support from the justice and fairness space.
We take a novel and different route in this paper: we look at the ideas of \emph{John Rawls}\footnote{\url{https://plato.stanford.edu/entries/rawls/}} -- an influential 20\textsuperscript{th} century moral and political philosopher in liberal tradition -- and attempt to incorporate his ideas of fairness in unsupervised ML.
Compared to other ideas, the Rawlsian ideas are time-tested, and a good mix of pragmatism and principledness, but incorporating them into ML algorithms is challenging.

In general, ideas of fairness can be incorporated in existing ML algorithms in 3 ways --
\begin{enumerate*}[label=(\roman*)]
    \item \emph{preprocessing}: where the input is processed to ensure fairness before being fed to the ML algorithm,
    \item \emph{inprocessing}: where the ML algorithm is altered to incorporate the fairness criteria (\Eg by adding fairness constraints to the optimisation criterion), and 
    \item \emph{postprocessing}: where the output of the ML algorithm is processed to make it fairer
\end{enumerate*}.
Preprocessing and postprocessing techniques can be used with any off-the-shelf ML algorithm with no modification, which is not the case with the inprocessing techniques.
Existing ML algorithms can thus be easily augmented to produce fairer outputs through preprocessing and postprocessing techniques.
On the other hand, the inprocessing techniques are known to have both better utility and better fairness than the other two.

In this paper, we focus on \emph{clustering} -- an unsupervised ML task.
We specifically look at the \emph{k-means clustering} algorithm \cite{hastie09the}, a well-known unsupervised ML algorithm that is used to partition a collection of examples into disjoint sets called \emph{clusters} such that similar examples are assigned to the same cluster.
We look at the k-means clustering algorithm with an additional constraint of satisfying Rawls' \emph{difference principle} \cite{rawls01justice}.
In other words, we allow the overall utility of the obtained clusters to be sub-optimal as long as the least-advantaged sensitive group has the greatest utility.
We work towards developing a \emph{postprocessing} technique that perturbs the output of the standard k-means clustering algorithm to incorporate the difference principle.
While there has been research on the intersection of fairness and clustering, to the best of our knowledge, this is the first to attempt to incorporate Rawlsian ideas in clustering.
Although this work is exploratory, we hope that this report of our experiences would aid future work in this direction.

\section{The Difference Principle and Rawlsian Point}
In his book \emph{Justice as Fairness: A Restatement}, John Rawls states the \emph{difference principle} as
\begin{displayquote}[\cite{rawls01justice}]
    \textquote{\emph{Social and economic inequalities are to satisfy two conditions: first, they are to be attached to offices and positions open to all under conditions of fair equality of opportunity; and second, they are to be to the greatest benefit of the least-advantaged members of society.}}
\end{displayquote}
We focus on the second condition: \textquote{\emph{they [social and economic inequalities] are to be to the greatest benefit of the least-advantaged members of society}}.
For the experiments outlined in this paper, we interpret the above statement as the members of a society belonging to one of two sensitive social groups (\Eg Male or Female)\footnote{We use binary genders for sake of illustration only.}, one being the more advantaged group and the other the less advantaged group.

\subsection{Terminology}
In the context of clustering, we refer to the \emph{Rawlsian point} as the cluster assignment where the difference principle is satisfied \Ie the utility to the least-advantaged members of society is the greatest, and we refer to such a cluster assignment as the \emph{Rawlsian k-means clusters}.
In contrast, the classical k-means clustering algorithm by design returns a cluster assignment where the sum of individual utilities is (approximately) maximised; we refer to this point as the \emph{utilitarian point}, and the corresponding cluster assignment as the \emph{utilitarian k-means clusters}.
This paper looks at binary sensitive attributes \Ie we assume that all examples in the dataset belong to one of two sensitive groups.
We refer to the sensitive group with the lower utility as the less advantaged group (LAG) and the one with the higher utility as the more advantaged group (MAG).
Thus, the LAG and MAG are defined on the sensitive attribute.
Note that the sensitive attribute is \emph{not} used in the k-means clustering algorithm.

\section{Problem statement\label{section:problem-statement}}
We attempt to develop a postprocessing technique that operates on the output of the k-means clustering algorithm, returning a cluster assignment that corresponds to the Rawlsian point \Ie the Rawlsian k-means clusters.
This requires that the Rawlsian point does indeed exist.
This leads us to the following questions:
\begin{enumerate}
    \item Does there exist a cluster assignment that corresponds to (or is an approximate of) the Rawlsian point? (Section \ref{section:existence})
    \item Can we reach this point starting from the utilitarian point \Ie the point of highest utility achieved by the classical k-means clustering algorithm? (Section \ref{section:algo})
\end{enumerate}

\subsection{Notation}
Let $\mathcal{X} = [ \ldots, x, \ldots ]$ be a collection of examples defined over a set of non-sensitive attributes $\mathcal{N}$ and a single binary sensitive attribute $S = \lbrace 0, 1 \rbrace$.
For any example $x$, let $x.n$ denote its values for the non-sensitive attributes and $x.s$ denote its value for the sensitive attribute.
Also, let $x.n \in [0, 1]^{|\mathcal{N}|}$ and $x.s \in \lbrace 0, 1 \rbrace$.
Note that the non-sensitive attributes $\mathcal{N}$ are the only attributes used for clustering; the clustering algorithm does not use the sensitive attribute $S$.

The k-means clustering algorithm assigns a label to each example based on the example's distance from the cluster centroids, thus yielding a cluster assignment $\mathcal{C} = \lbrace \ldots, C, \ldots \rbrace$ with $|\mathcal{C}| = k$, where each cluster $C$ is a set of examples having the same label and $k$ is the number of clusters to be generated.
We additionally define the \emph{utility} $u(x)$ of an example $x$ as
\begin{equation}
    u(x) = \delta - d(x.n, C)
\end{equation}
where the constant $\delta$ is the maximum possible distance between two examples, and $d(x.n, C)$ is the distance of example $x$ from the nearest cluster centroid.
Further, the utility of a sensitive group $\alpha$ is computed as
\begin{equation}
    U(\alpha) = \frac{1}{| \mathcal{X}_\alpha |} \sum_{x \in \mathcal{X}_\alpha} u(x)
\end{equation}
where $\mathcal{X}_\alpha$ denotes the set of examples in $\mathcal{X}$ belonging to the sensitive group $\alpha$ \Ie $\mathcal{X}_\alpha = \{x | x \in \mathcal{X}, x.s=\alpha \}$.
The overall utility of a cluster assignment is the average utility of all examples in the dataset.

In the next section we describe the Adult dataset which is central to this exploratory study, and in the subsequent sections we attempt to address the questions outlined above.

\section{Dataset}
We use the publicly available Adult dataset\footnote{\url{https://archive.ics.uci.edu/ml/datasets/adult}} from the UCI repository \cite{dua17uci} in our experiments.
The Adult dataset has been heavily used in the fairness literature.
It consists of 15 attributes and 30718 examples with no missing values.
We use 8 non-sensitive attributes, 1 sensitive attribute, and the predictor.
Table \ref{tab:adult-attributes} lists out the attributes from this dataset that are used for preprocessing, clustering, and evaluation in our experiments.
\begin{table}
    \centering
    \begin{tabular}{>{\bfseries}ll>{\ttfamily}l}
        \hline
                      & \textsc{type} & \textsc{name}  \\
        \hline
        Sensitive     & categorical   & sex            \\
        \hline
        Non-sensitive & continuous    & age            \\
                      &               & education-num  \\
                      &               & capital-gain   \\
                      &               & capital-loss   \\
                      &               & hours-per-week \\
        \cline{2-3}
                      & categorical   & workclass      \\
                      &               & education      \\
                      &               & occupation     \\
        \hline
        Predictor     & categorical   & annual-income  \\
        \hline
    \end{tabular}
    \caption{Attributes from the Adult dataset used in our experiments.
    Only the non-sensitive attributes are used in the clustering algorithm.
    The sensitive attribute is used only for computing the utilities of the sensitive groups.
    The predictor attribute is used only when preprocessing the dataset.
    \label{tab:adult-attributes}}
\end{table}

\subsubsection*{Preprocessing}
We follow the steps in previous work \cite{abraham20fairness} for preprocessing the dataset.
The non-sensitive attributes (used for clustering) are preprocessed as follows:
\begin{itemize}
    \item continuous: scaled and translated to the range $[0, 1]$.
    \item categorical: one hot encoded. To ensure that the maximum squared distance between any two examples for this attribute is 1, the value for the `hot' position is set to $\frac{1}{\sqrt{2}}$.
\end{itemize}
Consequently, the maximum distance between any two values of a single non-sensitive attribute is 1.
As we use 8 non-sensitive attributes, the maximum possible distance $\delta$ between any two examples is 8.
We then undersample for parity across the predictor attribute.
The resulting dataset contains 42 non-sensitive attributes and 1 sensitive attribute.
Finally, we sample 500 examples from each predictor class for a total of 1000 examples.
Table \ref{tab:adult-sensitive} shows the distribution of sensitive groups in the sampled dataset.
\begin{table}
    \centering
    \begin{tabular}{lrr}
        \hline
        sex    & \# of examples & \% of examples \\
        \hline
        Female &            267 &         26.7\% \\
        Male   &            733 &         73.3\% \\
        \hline
    \end{tabular}
    \caption{Distribution of sensitive groups (Female, Male) in the sampled dataset (1000 examples).
    \label{tab:adult-sensitive}}
\end{table}

\section{Existence of Rawlsian k-means clusters\label{section:existence}}
\begin{figure}[ht]
    \centering
    \begin{subfigure}[t]{.48\textwidth}
        \centering
        \includegraphics[scale=.49]{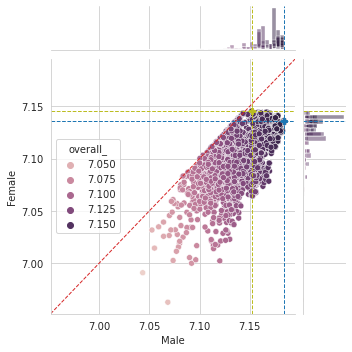}
        \caption{points generated by the 5000 runs}
    \end{subfigure}%
    \quad
    \begin{subfigure}[t]{.48\textwidth}
        \centering
        \includegraphics[scale=.49]{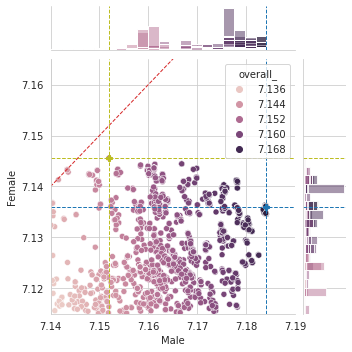}
        \caption{zooming in to the utilitarian point and (approximate) Rawlsian point}
    \end{subfigure}
    \caption{Scatter plot of points in the MAG-LAG utility space generated by 5000 runs of the k-means clustering algorithm with different initial centroids, $k=5$, on our Adult dataset, and majority (Male) as the more advantaged group (MAG) and minority (Female) as the less advantaged group (LAG).
    Each point in this space corresponds to the cluster assignment obtained from a single run.
    The hue of a point indicates the overall utility, with darker being better.
    Points on the 45° red dashed line are such that the utilities for MAG and LAG are equal.
    By design of this experiment \Ie the minority group (Female) being the less advantaged group, all points are always south of this 45° line.
    The horizontal/vertical olive dashed lines intersect at the point with the highest LAG (Female) utility, but has low overall utility since Female is a minority group (see Table \ref{tab:adult-sensitive}).
    This olive point is an approximate for the Rawlsian point.
    The horizontal/vertical blue dashed lines intersect at the point with the highest MAG (Male) utility, and is also of the highest overall utility (\Ie the utilitarian point) since Male is a majority group.
    The histograms to the top and right of each scatter plot indicate the density of the points.
    \label{fig:existence}}
\end{figure}

Before devising a technique for obtaining the Rawlsian k-means clusters, we need to determine whether such a cluster assignment indeed exists.
Since it is impractical to enumerate and evaluate the utilities of all possible cluster assignments, we instead perform several runs of the k-means algorithm on our dataset with different initial centroids to find an approximate.
Note that either of the sensitive groups may be less (or more) advantaged; in case of the Adult dataset, we only consider those cluster assignments where the minority group (\Ie Female) is the less advantaged group.
Figure \ref{fig:existence} shows the utilities of these cluster assignments, with the utility of the more advantaged group (MAG) on the $x$-axis and the less advantaged group (LAG) on the $y$-axis.

It can be seen in Figure \ref{fig:existence} that there is indeed a point (shown as an olive plus point) corresponding to a k-means clustering that has a better utility for the less advantaged group than the utilitarian point (shown as a blue plus point).
The olive plus point is thus an approximate for the Rawlsian point.
Note that this point may not be the actual Rawlsian point; we use it as an approximate for the Rawlsian point in the experiment discussed in Section \ref{section:algo} as evaluating all possible cluster assignments is not possible.

\section{Reaching the approximate Rawlsian point\label{section:algo}}
We saw in the previous section there is no mechanism to generate points in the MAG-LAG utility space other than through the k-means algorithm itself.
We thus performed several runs of the k-means algorithm to see whether doing so can reveal spaces where the Rawlsian point is likely to be and whether we can navigate to these spaces.
Now, the histograms to the top and right of the scatter plots in Figure \ref{fig:existence}, indicate that the generated points are concentrated near the utilitarian point (the blue plus point).
This would suggest that a k-means algorithm would likely generate a cluster assignment that corresponds to a point closer to the utilitarian point.
Thus, we select the utilitarian cluster assignment \Ie the one with highest overall utility (shown as a blue plus point in Figure \ref{fig:existence}) as the starting point for our postprocessing technique.

We now outline our postprocessing technique for arriving at the approximate Rawlsian point starting from the utilitarian point.
Our goal is to find a \emph{reassignment} of examples to new clusters so that we reach the approximate Rawlsian point (shown as a olive plus point).
To do so, we apply a series of reassignment operations to the utilitarian cluster assignment such that we gradually move above the horizontal blue line in the MAG-LAG utility space toward the Rawlsian point.
The general outline of our technique is shown in Algorithm \ref{algo:traverse}.
\begin{algorithm}[ht]
    \caption{Traverse \label{algo:traverse}}
    \begin{algorithmic}[1]
        \Require $\mathcal{X}$, $\mathcal{C}$
        \Ensure $\mathcal{C}$
        \While{$\mathcal{C}$ has not changed}
            \State $O \gets$ GenerateOperations$(\mathcal{X}, \mathcal{C})$
            \Comment{Generate operations (Section \ref{section:operators})}
            \State $o \gets$ select the best operation in $O$
            \Comment{(Section \ref{section:select-apply})}
            \If{$o$ is not $\phi$}
                \State $\mathcal{C} \gets$ apply the selected operation $o$
            \EndIf
        \EndWhile
    \end{algorithmic}
\end{algorithm}
We discuss the GenerateOperations algorithm in Section \ref{section:operators}, and how the best operation is selected and applied in Section \ref{section:select-apply}.

\subsection{Selecting and applying the best reassignment operation\label{section:select-apply}}
We select the best reassignment operation among those generated according to the following order of preference:
\begin{enumerate}
    \item Among the operations that generate a point in the north-east of the current point in the MAG-LAG utility space, select the operation that corresponds to the point with highest overall utility.
    \item If no such operation exists, among the operations that generate a point in the \emph{skyline} of points in the north-west of the current point, select the operation that corresponds to the point with highest overall utility.
    \item If no such operation exists, select the null operation $\phi$, which indicates that no reassignment is done and hence the cluster assignment is unchanged.
\end{enumerate}
The \emph{skyline} $S$ of a set of points $Q$ is defined as those points in the MAG-LAG utility space where $\forall q \in$ Q, $\exists s \in S$ such that either
\begin{enumerate*}[label=(\roman*)]
    \item $q == s$ (\Ie $q$ is in the skyline), or
    \item $u_\mathit{LAG}(q) < u_\mathit{LAG}(s)$ and $u_\mathit{MAG}(q) < u_\mathit{MAG}(s)$ (\Ie $q$ is worse than $s$ for both LAG and MAG), where $u_\alpha(x)$ is the utility of sensitive group $\alpha$ for the point $x$
\end{enumerate*}.
Moving through the skyline ensures that we select the operation with the least drop in overall utility and maximum gain in LAG utility.

Applying the selected reassignment operation is straightforward; we change the current cluster assignment (\Ie current labels of the examples) as specified by the selected operation, and use the new cluster assignment as the starting point for the next iteration.

\section{Reassignment operators\label{section:operators}}
Our goal is to construct an operator that reassigns a number of examples in the current cluster assignment to new clusters, thus generating a new cluster assignment -- the Rawlsian cluster assignment -- having a higher LAG utility while minimally affecting the overall utility.
By applying a series of instantiations of this operator to the utilitarian cluster assignment, we hope to reach the approximate Rawlsian point.
We explore two simple operators $\Rone$ and $\Rtwo$ which are now detailed.

\subsection{Reassignment operator $\Rone$\label{section:r1}}
The reassignment operator $\Rone$ operating on a tuple $(x, C')$ takes a single example $x$ from the current cluster assignment $\mathcal{C}$ and reassigns it to a different cluster $C'$ (\Ie a different label) thus yielding a new cluster assignment.
The number of possible operations\footnote{An \emph{operation} is an instantiated operator.} generated for $\Rone$ is thus $n \times (k-1)$.
Algorithm \ref{algo:generate-r1} outlines the GenerateOperationsR1 algorithm which is the $\Rone$ variant of the GenerateOperations algorithm.
\begin{algorithm}[ht]
    \caption{GenerateOperationsR1\label{algo:generate-r1}}
    \begin{algorithmic}[1]
        \Require $\mathcal{X}$, $\mathcal{C}$
        \Ensure $O$
        \State Initialise $O$ to an empty set
        \For{each example-cluster tuple $(x, C') \in \mathcal{X} \times \mathcal{C}$}
            \State Apply $\Rone(x, C')$ to get a new cluster assignment $\mathcal{C}'$
            \State Obtain the corresponding point in the MAG-LAG utility space
            \If{the point has a higher LAG utility than the current point}
                \State Add $(x, C')$ to $O$
            \EndIf
            \State Discard $\mathcal{C}'$
        \EndFor
    \end{algorithmic}
\end{algorithm}

\begin{figure}
    \centering
    \includegraphics[scale=.7]{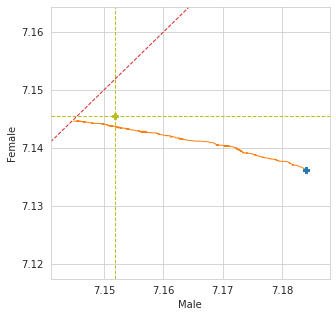}
    \caption{Trajectory of points (in orange) generated in the MAG-LAG utility space by Traverse (Algorithm \ref{algo:traverse}) using the $\Rone$ operator (Section \ref{section:r1}).
    Points on the 45° red dashed line are such that the utilities for MAG and LAG are equal.
    The olive plus point corresponds to the approximate Rawlsian point which we would like the Traverse algorithm to arrive at.
    Recall that the Rawlsian point corresponds to the cluster assignment where the utility of the less advantaged group is the greatest.
    The algorithm starts from the utilitarian point (indicated by the blue plus point) and repeatedly applies a $\Rone$ operation thus yielding a new cluster assignment, with each cluster assignment generating a corresponding point in the utility space.
    This figure contains 261 such generated points.
    The orange line that connects the generated sequence of points together indicates the trajectory of the algorithm which is generally moving in the north-west direction, and can be seen to terminate at the red line \Ie where the utilities for MAG and LAG are equal.\label{fig:traverse-r1}}
\end{figure}

Figure \ref{fig:traverse-r1} shows the trajectory of points in the MAG-LAG utility space generated by employing GenerateOperationsR1 in Algorithm \ref{algo:traverse}.
We see that the algorithm takes small steps (each step corresponds to the application of one $\Rone$ operation) in the correct direction towards the approximate Rawlsian point but ends up with the LAG and MAG utilities being equal.
Notably, the approximate Rawlsian point (olive plus point in the figure) has a better LAG (Female) utility than all generated points.
There is thus a need for improvement.
Another drawback of $\Rone$ is that its repeated application may be inefficient because the amount of movement towards the approximate Rawlsian point in each iteration is negligible.
Additionally, the reassignment operation may never be undone.

\subsection{Pair reassignment operator $\Rtwo$}
To overcome the inefficiency limitation of $\Rone$, we instead select pairs of examples to be reassigned.
We define the \emph{pair reassignment} operator as follows:
The pair reassignment operator $\Rtwo$ operating on a pair of tuples $(x_1, C'_1), (x_2, C'_2)$ takes a pair of distinct examples $x_1$ and $x_2$ ($x_1 \ne x_2$) from the current cluster assignment $\mathcal{C}$ and reassigns them to new clusters $C'_1$ and $C'_2$ thus yielding a new cluster assignment.
One $\Rtwo$ operation is thus equivalent to two $\Rone$ operations on two distinct examples $x_1$ and $x_2$.

Two issues arise when using $\Rtwo$ for navigating the MAG-LAG utility space:
\begin{enumerate}
    \item The number of possible operations using $\Rtwo$ is ${n \choose 2} (k-1)^2$, where $n$ is the number of examples and $k$ is the number of clusters.
    In our experiment, this evaluates to nearly 8 million possible operations; it is impractical to calculate the utilities of all such operations.
    In contrast, the number of possible operations using $\Rone$ is $n \times (k-1)$ which is significantly smaller.
    \item On inspection of a sample of these $\Rtwo$ operations, we found that a large chunk (around 62\%) generate points with lower LAG utilities than the current point \Ie they move away from the Rawlsian point, and hence are not useful.
\end{enumerate}
Thus, there is a need to intelligently generate $\Rtwo$ operations whose corresponding points have higher LAG utilities than the current point.
Algorithm \ref{algo:generate-r2} outlines the GenerateOperationsR2 algorithm which is the $\Rtwo$ variant of the GenerateOperations algorithm.
Instead of generating all possible operations, we use a heuristic for pruning the set of example-cluster tuples (Lines 1 to 4 in Algorithm \ref{algo:generate-r2}).
This heuristic is based on the assumption that if $\Rone$ does not generate a point with higher LAG utility for some tuple $(x, C')$, then no $\Rtwo$ operation instantiated with $(x, C')$ -- \Ie neither $\Rtwo(x, C', \cdot, \cdot)$ nor $\Rtwo(\cdot, \cdot, x, C')$ -- will generate a point with higher LAG utility than the current point.
Next, the remaining tuples are separately ranked on LAG utility and overall utility, and only the top reassignments (1\% to 5\%) from the two rankings are retained.
This yields a set of good example-cluster tuples $T_\mathit{good} \subset \mathcal{X} \times \mathcal{C}$ that can be later used to instantiate $\Rtwo$.
The rest of the algorithm is similar to Algorithm \ref{algo:generate-r1}.
\begin{algorithm}[ht]
    \caption{GenerateOperationsR2\label{algo:generate-r2}}
    \begin{algorithmic}[1]
        \Require $\mathcal{X}$, $\mathcal{C}$
        \Ensure $O$
        \State $T \gets$ GenerateOperationsR1$(\mathcal{X}, \mathcal{C})$
        \State $T_\mathit{LAG} \gets$ top $p$\% of $T$ ranked on LAG utility
        \State $T_\mathit{overall} \gets$ top $q$\% of $T$ ranked on overall utility
        \State $T_\mathit{good} \gets T_\mathit{LAG} \cup T_\mathit{overall}$
        \State Initialise $O$ to an empty set
        \For{each pair of example-cluster tuples $((x_1, C'_1), (x_2, C'_2)) \in T_\mathit{good} \times T_\mathit{good}$}
            \State Apply $\Rtwo(x_1, C'_1, x_2, C'_2)$ to get a new cluster assignment $\mathcal{C}'$
            \State Obtain the corresponding point in the MAG-LAG utility space
            \If{the point has a higher LAG utility than the current point}
                \State Add $(x_1, C'_1, x_2, C'_2)$ to $O$
            \EndIf
            \State Discard $\mathcal{C}'$
        \EndFor
    \end{algorithmic}
\end{algorithm}

\begin{figure}
    \centering
    \includegraphics[scale=.7]{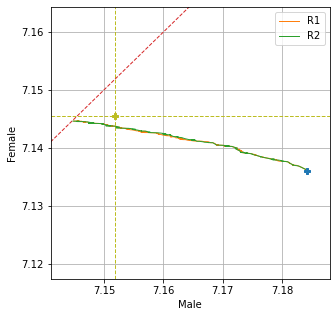}
    \caption{Trajectories of points generated in the MAG-LAG utility space by Traverse (Algorithm \ref{algo:traverse}) using the $\Rone$ operator (in orange) and the $\Rtwo$ operator (in green).
    Points on the 45° red dashed line are such that the utilities for MAG and LAG are equal.
    The olive plus point corresponds to the approximate Rawlsian point which we would like the Traverse algorithm to arrive at.
    Recall that the Rawlsian point corresponds to the cluster assignment where the utility of the less advantaged group is the greatest.
    The algorithm starts from the utilitarian point (indicated by the blue plus point) and repeatedly applies a $\Rone$ (or $\Rtwo$) operation thus yielding a new cluster assignment, with each cluster assignment generating a corresponding point in the utility space.
    $\Rone$'s trajectory consists of 261 such generated points, and $\Rtwo$'s trajectory consists of 152 such generated points.
    Similar to $\Rone$'s trajectory, $\Rtwo$ moves in the north-west direction terminating at the red line \Ie where the utilities for MAG and LAG are equal.\label{fig:traverse-r1vr2}}
\end{figure}

Figure \ref{fig:traverse-r1vr2} compares the trajectories of points in the MAG-LAG utility space generated by employing GenerateOperationsR1 and GenerateOperationsR2 in Algorithm \ref{algo:traverse}.
We can see that while $\Rtwo$ improves upon the inefficiency limitation of $\Rone$ (requiring 152 iterations as compared to 261 iterations by $\Rone$), it follows nearly an identical trajectory as $\Rone$ and hence still suffers from its other limitations \Ie ends up with the LAG and MAG utilities being equal, and the operation once applied may never be undone.

\section{Related Work}
While there has been significant research on fairness in machine learning in recent years, we look at those works that are relevant to this paper \Ie
\begin{enumerate*}[label=(\roman*)]
    \item Rawlsian ideas of fairness in machine learning, and
    \item fair algorithms for clustering.
\end{enumerate*}.

\textbf{Rawlsian ideas of fairness in ML:}
The few existing works that explore Rawls' ideas of fairness in ML are in the supervised setting.
The hardness of adapting Rawlsian principles into algorithms is apparent from these works.
For example, Shah et al \cite{shah21rawlsian} propose a classifier that minimises the error rate of the worst-off sensitive group; they call this a Rawls classifier.
Hashimoto et al \cite{hashimoto18fairness} employ Rawlsian ideas to mitigate the amplification of representation disparity in empirical risk minimization.
We have not come across any work that explores Rawls' ideas in the unsupervised setting.

\textbf{Fair algorithms for clustering:}
These can be broadly categorised based on the notion of fairness \Ie
\begin{enumerate*}[label=(\roman*)]
    \item individual fairness \cite{jung19a,kleindessner20a,mahabadi20individual}, and
    \item group fairness \cite{chierichetti17fair,davidson20making,kleindessner19fair}
\end{enumerate*}.
Further, these algorithms differ on how the fairness criterion is enforced (\Ie preprocessing, inprocessing, or postprocessing).
For example, Chierichetti et al \cite{chierichetti17fair} propose a preprocessing technique that makes the output of a subsequent standard clustering algorithm fair.
Kleindessner et al \cite{kleindessner19fair} incorporates the fairness constraint within the clustering algorithm.
Davidson and Ravi \cite{davidson20making} look at postprocessing clusters for fairness; they do so by presenting it as a Minimum Cluster Modification for Group Fairness (MCMF) optimisation problem which is formulated as an ILP.
Other notions of fairness such as representativity fairness \cite{p20representativity} and proportionality fairness \cite{chen19proportionally} have also been proposed for clustering.

\section{Conclusion}
We proposed a postprocessing framework for making the clusters generated by the standard k-means clustering algorithm satisfy Rawls' difference principle while minimally affecting the overall utility.
Within this framework, we explored two simple operators that perturb a given cluster assignment by reassigning examples to new clusters; the first operator $\Rone$ reassigns a single example to a new cluster at a time, and the second operator $\Rtwo$ reassigns two examples to new clusters at a time.
We observed that while $\Rtwo$ improves upon the efficiency limitation of $\Rone$, there is still a huge scope for improvement with regards to arriving at the Rawlsian point.
There is a need to design an operator that reassigns a larger number of examples to other clusters, which will consequently open up new avenues for exploration in the search space.
Nevertheless, we expect these operators to act as good baselines for any future operator.

\section*{Acknowledgment}
This project has received funding from the European Union’s Horizon 2020 research and innovation programme under the Marie Skłodowska-Curie grant agreement No 945231; and the Department of the Economy in Northern Ireland.

%
%
%
\bibliographystyle{splncs04}
\bibliography{mybibliography}
\end{document}